\documentclass[conference]{IEEEtran}
\usepackage{array}
\usepackage{booktabs}
\usepackage{graphicx}
\usepackage{color}
\usepackage[ruled,vlined]{algorithm2e}
\usepackage{cite}
\usepackage{amsmath}
\usepackage{algorithmic}
\begin{document}
\title{ARC-Net: Activity Recognition Through Capsules}
\author{
\IEEEauthorblockN{Hamed Damirchi, Rooholla Khorrambakht and
Hamid D. Taghirad, Senior Member, IEEE}
\IEEEauthorblockA{Faculty of Electrical and Computer Engineering\\
    K. N. Toosi University of Technology\\
    P.O. Box 16315-1355, Tehran, lran\\
    Email: hdamirchi, r.khorrambakht@email.kntu.ac.ir, taghirad@kntu.ac.ir}}

\maketitle

\begin{abstract}
Human Activity Recognition (HAR) is a challenging problem that needs
advanced solutions than using handcrafted features to achieve a
desirable performance. Deep learning has been proposed as a solution
to obtain more accurate HAR systems being robust against noise. In
this paper, we introduce ARC-Net and propose the utilization of
capsules to fuse the information from multiple inertial measurement
units (IMUs) to predict the activity performed by the subject. We
hypothesize that this network will be able to tune out the
unnecessary information and will be able to make more accurate
decisions through the iterative mechanism embedded in capsule
networks. We provide heatmaps of the priors, learned by the network,
to visualize the utilization of each of the data sources by the
trained network. By using the proposed network, we were able to
increase the accuracy of the state-of-the-art approaches by 2\%.
Furthermore, we investigate the directionality of the confusion
matrices of our results and discuss the specificity of the
activities based on the provided data.

\end{abstract}


%
\IEEEpeerreviewmaketitle

\section{Introduction}

In the human activity recognition field, the goal is to predict the
activity of a human subject based on a window of measurements
provided by the available sensors. These activities may range from
lying to rope jumping. A wide variety of sensors such as
accelerometer, gyroscope, magnetometer, force sensor, and light
sensor may be used to classify the activity performed in that window
of collected data. Even everyday devices such as smartphones and
smartwatches may be used in order to obtain the required data for
activity recognition. Due to availability and the low cost of the
mentioned sensors, HAR is utilized in many areas such as design of
exoskeletons~\cite{exo} and more commonly, elderly
care~\cite{harhealthcare}.  Furthermore, the collected data from HAR
sensors may be collected and utilized to provide more personalized
services through the monitored activities. For example, the data
collected from body-worn sensors and the behavior extracted form
this data can help increase security~\cite{harsecurity}, provide
more modern and reactive healthcare~\cite{harhealthcare} and allow
the creation of better user interfaces~\cite{harUI}.

Deep learning has proven to be an adequate tool for recognizing
patterns and extracting rich features that may be utilized to
classify data~\cite{classificationdeep}. Human
activities may be classified based on the patterns that are seen in
the input data. Therefore, rather than looking only at the
measurements from individual data sources, we require our algorithm
to view the input data on different levels and fuse the extracted
features in such a way that the perceived patterns would be able to
tell us about the activity being performed, e.g., to differentiate
between walking and running, it would be misleading to only look at
the data provided by a motion sensor located on the ankle of the
subject and we would reach better performances by also using the
data from subject's forearm. This calls for a technique that is able
to adequately fuse the information from every data source and use
the said information to make a prediction about the activity.

In this paper,  we propose using a single Convolutional Neural
Network (CNN) as an encoder to extract the features from each of the
IMUs and pass the said features to CapsNet~\cite{capsnet} in order
to fuse the extracted information and make a prediction about the
activity of the subject. We will use two datasets to evaluate the
performance and generalization of our approach. Moreover, a modified
version of the network proposed in~\cite{PNet} will be used as our
encoder which uses stage-based fusion to extract the necessary
information. Our contributions are as follows:

\begin{itemize}
    \item CapsNet is used to fuse the information obtained from each IMU
    \item Provide intuitive interpretations regarding the
     utilization of each of the IMUs based on the true label
    \item Provide comparisons to empirically examine the capability
    of capsules in rejecting corrupted modalities
    \item Our method outperforms the state-of-the-art deep
    learning based approaches
\end{itemize}

This paper is organized as follows: In Section 2, we go through
various available classical and deep learning-based approaches to
HAR alongside the current state-of-the-art(SOTA) approaches. In
Section 3, we explain the proposed architecture and provide an
introduction to capsules. In the last section, we present
quantitative and qualitative evaluations of the proposed method
against SOTA approaches and provide visualizations alongside
corresponding discussions.

\section{Related work}
From an algorithmic point of view, approaches to HAR may be divided
into three groups, namely, classical approaches that utilize
preprocessing methods such as pose estimation, machine
learning-based approaches that rely on hand-crafted features devised
by an expert and deep learning-based methods that rely on gradient
backpropagation in order to both extract features and perform
classification. Deep learning based methods can be separated into
multiple groups of architectures themselves. Each group utilizes a
specific characteristic of an architecture to improve the accuracy
of their predictions. This category of approaches may be divided
further based on the memory of past inputs~\cite{DeepConvLSTM,DCR,rnnmedical}, unsupervised methods~\cite{deepautoset, unsupervisedhar},
non-recurrent methods~\cite{HARSmartphone,PNet}.

In~\cite{fasterSVM}, video sequences captured by a monocular camera
are used to classify the activity of subjects. Optical flow
alongside background extraction methods is used as features which
are then passed to a support vector machine (SVM) to generate
predictions. \cite{decisiontree} uses decision trees to classify the
activity of the subject based on features extracted from an
accelerometer of a smartphone. Ten features such as phone position
on the human body, user location, age and sensor readings are passed
through a decision tree to make predictions. ~\cite{deepautoset}
uses activity sets rather than single label ground truth values to
predict the potential activities in the corresponding window of
measurements from IMUs. This work also uses unsupervised learning
methods prior to supervised learning in their training process to
achieve a more effective feature representation.
In~\cite{unsupervisedhar}, unsupervised learning methods are
deployed where the number of activities is unknown. This is done by
using clustering methods that operate on the frequency components of
the measured acceleration and angular velocity values.

\cite{DeepConvLSTM} proposes DeepConvLSTM which is an LSTM based
network that utilizes a CNN to extract the features from inputs. Due
to the usage of recurrent layers, this model falls into the category
of memory-based networks. DeepConvLSTM aims to increase the
performance of HAR systems by modeling temporal dependencies while
using raw sensor data. In~\cite{DCR} various architecture choices
are compared to reach a conclusion about the necessity of recurrence
in HAR models. \cite{rnnmedical} uses variants of Recurrent Neural
Networks (RNNs) to observe the cognitive decline by monitoring the
daily activities of the subject, while~\cite{DeepConvLSTM, DCR,
rnnmedical} all use IMU measurements as network input.

\cite{HARSmartphone} stacks IMU measurements from a smartphone and
creates a window of measurements. These measurements are then passed
through a CNN to predict the activity of the subject. This work aims
to model the temporal connections between raw sensor measurements
using the convolution operations that reside in CNNs. \cite{PNet}
also stacks the raw measurements from the available IMUs and uses
convolutional layers to extract the features from said inputs. This
work uses specific kernel sizes that allow the network to perform
data fusion in multiple levels. Reference~\cite{PNet} also proposes
that late sensor fusion is more effective due to the separate
processing of each axis of sensor module in the initial layers. We
base our encoder on this architecture with modifications to prevent
loss of information during pooling layers. Moreover, we view HAR as a multimodal fusion problem. We use a single encoder to extract features from IMUs with no constraints on the position of the data source on subject's body. Thereafter, we use mechanisms that fuse the collected information and predict the performed activity.
\begin{figure}[t]
    \hspace*{0.5cm}
    \includegraphics[scale=0.76]{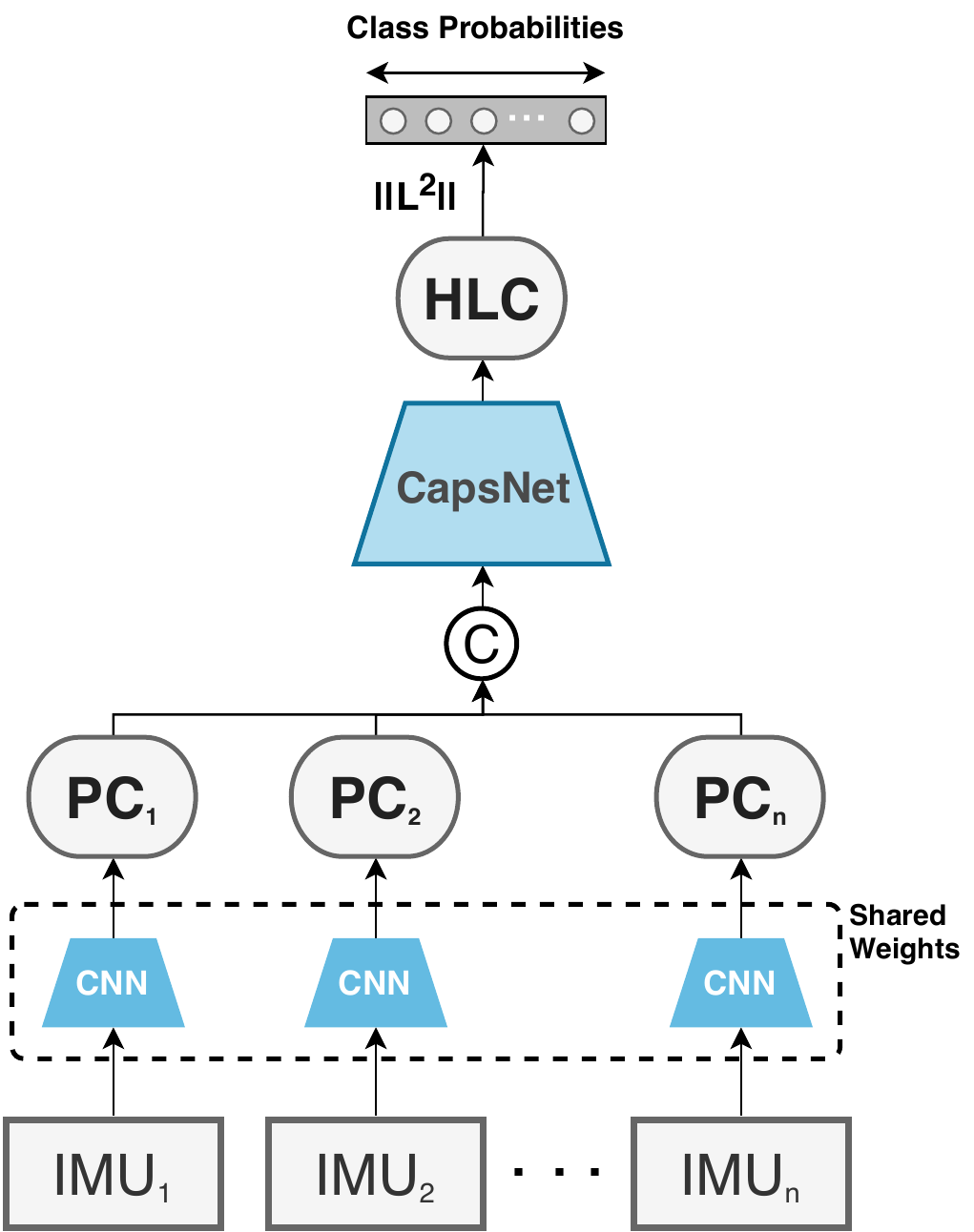}
    \caption{A general overview of the proposed architecture}
    \label{architecture}
\end{figure}

\section{The Proposed Approach}
We propose using CapsNets  to fuse the high dimensional features
passed from an encoder. The general overview of our architecture is
shown in Fig. \ref{architecture}. We stack a pre-specified number of
measurements from each IMU and create a two-dimensional array where
columns represent each measurement. These arrays are passed to a
single CNN separately and the features corresponding to each of the
IMUs are extracted. Then, primary capsules are formed by reshaping
the output of the CNN and concatenating the extracted features from
each of the encoders. The concatenated features are then passed
through CapsNet to obtain a probability vector of the activities.
These steps are elaborated in detail in Section 3.A and 3.B.
\begin{figure*}[t]
    \hspace*{0.2cm}
    \includegraphics[scale=0.78]{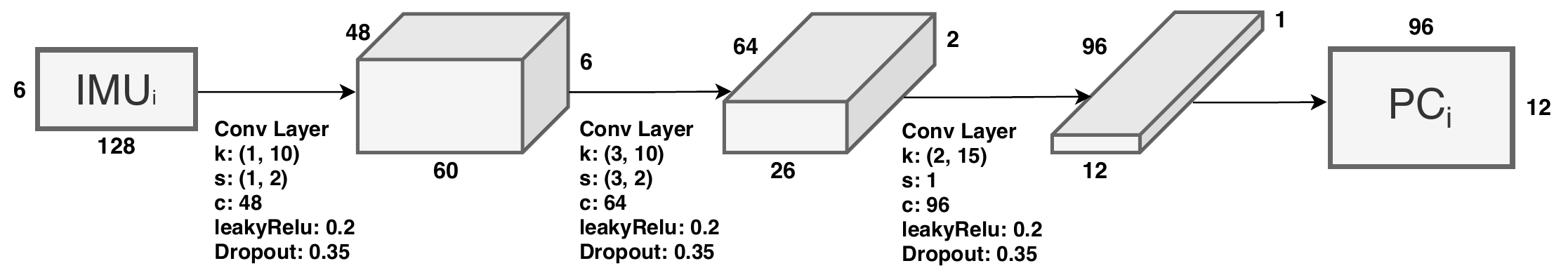}
    \caption{The architecture of the encoder. k, s and c correspond to kernel size, stride size and number of output channels, respectively}
    \label{encoder}
\end{figure*}

\subsection{Encoder}
In order to extract features from each of the IMUs, we use a CNN
with varying kernel sizes at each layer similar to the architecture
proposed in~\cite{PNet}. A modified version of this architecture is
presented in Fig.~\ref{encoder}. As it is seen in this figure, at
the first layer of the encoder, we perform a 1-dimensional
convolution operation that does not perform any fusion on the given
inputs and mainly acts as a filter. In the next layer, a
2-dimensional convolution is applied to the features of the former
layer. Due to the size of the kernel and the strides of this layer,
each module of the IMU is processed separately and the features of
the accelerometer and gyroscope are not fused in this layer. In the
final layer, a kernel size of ($2\times15$) convolves over the
features of the second layer and fuses the information extracted
from each of the modules. To be able to pass the extracted features
to CapsNet, we will need to reshape these features as shown in
Fig.~\ref{encoder}.

\subsection{CapsNet and Dynamic Routing}
CapsNet~\cite{capsnet} uses vectors of neurons (capsules) to
represent the entities that may be present in the given input. Each
capsule looks at a small window of the input and gives the
probability of the existence of a specific pattern in the data. The
magnitude of a capsule gives the probability of the existence of an
entity and the orientation of the capsule in its high dimensional
space defines the characteristics of that entity. The capsules in
the first layer of the network, which  are called primary capsules,
are extracted through the encoder as shown in Fig.~\ref{encoder}.
CapsNet uses an iterative mechanism to dynamically route each of the
lower layer capsules to the higher layer ones. Through this
mechanism, the network is able to use the information about the
existence of low level entities to decide about the presence of
higher level ones. In our approach we only use one layer on top of
the primary capsules in order to obtain predictions about the
activity performed by the subject. Therefore, the higher level
capsules will represent the activity that is potentially present in
the given window of measurements. In other words, the routing
mechanism will allow us to link the existence of specific patterns
in each of the IMU measurements to the potential activity that is
being performed by the subject.

By concatenating the primary capsules from each of the IMUs, we will
have a feature map of size $12n\times96$ where $n$ is the number of
IMUs. As seen in Fig. \ref{encoder}, each primary capsule has 96
dimensions and 12 capsules are extracted from each of the IMUs.
Assuming $U_i$ represents the $i^{th}$ lower layer capsule and $V_j$
is the $j^{th}$ higher layer capsule, Algorithm~\ref{capsnetalgo}
describes the routing mechanism used in our approach for r number of
iterations. The squash non-linearity is formulated as follows
\begin{equation}
\textrm{squash}(\mathbf{\hat{V}}_{j})=\frac{\left\|\mathbf{\hat{V}}_{j}
\right\|^{2}}{1+\left\|\mathbf{\hat{V}}_{j}\right\|^{2}}\cdot
\frac{\mathbf{\hat{V}}_{j}}{\left\|\mathbf{\hat{V}}_{j}\right\|}
\end{equation}
While, the softmax function is formulated as follolws:
\begin{equation}
\textrm{softmax}(b)=\frac{\exp(b_{ij})}{\sum_{k}\exp(b_{ik})}
\end{equation}
Following ~\cite{learnedrepcaps}, we added a soft updating rule with
the coefficient $\eta$ in Algorithm~\ref{capsnetalgo} to prevent
overrouting. Furthermore, we used the original loss function
from~\cite{capsnet} which is a margin loss that calculates a
separate loss value for each of the predicted classes.

\begin{algorithm}[t]
    \SetAlgoLined
    initialize the log prior matrix b and set $b_{ij}\leftarrow0$\\
    $\hat{U}_{j|i}$ = $U_iW_{ij}$\\
    \vspace{0.07cm}
    $c_{ij}\leftarrow \textrm{softmax}(b)$\\
    \For{r iterations}{
        $\hat{c}_{ij}\leftarrow \textrm{softmax}(b)$\\
        \vspace{0.07cm}
        $c_{ij}\leftarrow\eta\hat{c}_{ij}+c_{ij}$\\
        \vspace{0.07cm}
        $\hat{V}_j\leftarrow\sum_{i}c_{ij}\hat{U}_{j|i}$\\
        \vspace{0.07cm}
        $V_j\leftarrow \textrm{squash}(\hat{V}_j)$\\
        $b_{ij}\leftarrow b_{ij}+\hat{V}_j\cdot\hat{U}_{j|i}$\\
    }
    \caption{Dynamic routing algorithm}
    \label{capsnetalgo}
\end{algorithm}

\section{Experiments}
The experiments were conducted using an NVIDIA Tesla P100 with 16
gigabytes of RAM and 3584 CUDA cores. We used the
PyTorch~\cite{pytorch} framework to implement the proposed
architecture and PyTorch lightning~\cite{lightning} as an interface for
reproducibility purposes.

\subsection{Datasets}
The proposed method was tested on the PAMAP2~\cite{pamap2} and
RealWorld~\cite{realworld} HAR datasets. The validation set of the
PAMAP2 dataset was used to tune the model hyperparameters. The
quantitative results from each of the datasets were compared against
the state of the art methods with the same preprocessing
characteristics. We report our results against
PerceptionNet~\cite{PNet}, DeepConvLSTM~\cite{DeepConvLSTM} and
CNN-EF~\cite{HARSmartphone} on the PAMAP2 dataset and to test the
generalizability of our method, we also provide quantitative results
for PerceptionNet~\cite{PNet} and DeepConvLSTM~\cite{DeepConvLSTM}
alongside our method on the RealWorld dataset. To train the network
and to infer the activity of each of the subjects from either
dataset, we only use the IMU measurements, namely accelerometer and
gyroscope.

\subsubsection{RealWorld}
This dataset contains measurements from 15 subjects (8 males and 7
females) with recorded data from the chest, forearm, head, shin,
thigh, upper arm, and waist of each of the subjects. For each
subject, this dataset provides measurements from GPS, IMU,
gyroscope, light, sound level data and magnetic field sensors. The
annotated activities from this dataset are climbing up and down the
stairs, jumping, lying, standing, sitting and running. The data is
not synchronized and is recorded at a frequency of 50Hz. After
synchronizing the IMUs, we segment the data into windows of 128
measurements (2.56 seconds) with 60\% overlap. We use the
leave-one-subject-out approach to evaluate our method. Specifically,
we use subjects 10 and 11 as validation and test subjects and the
rest are used to train the network.

\subsubsection{PAMAP2}
The physical activity monitoring dataset consists of 18 different
physical activities from 9 subjects (8 males and 1 female). Three
Colibri wireless inertial measurement units were used to provide
measurements alongside a heart rate monitor. This dataset contains
IMU measurements from the chest, dominant wrist and dominant ankle
at a frequency of 100Hz. We downsampled this dataset to 50Hz in
order to match the frequency of the RealWorld dataset and stacked
128 measurements with the same overlapping that was used for the
preprocessing stage of the RealWorld dataset. Similar to
PerceptionNet, we chose a leave-one-subject-out approach to validate
and test our model. Subjects 1 and 5 were chosen as test and
validation sets, respectively.

\subsection{Performance evaluation metrics}
Due to an imbalance in the number of labels for both of the
datasets, we chose the weighted $F1$ score to report the
quantitative results of our network. This score is formulated as
below
\begin{equation}
    wF1=\sum_{c}\frac{N_c}{N}\frac{2 \cdot \textrm{Precision}_c\cdot \textrm{Recall}_c}
    {\textrm{Precision}_c + \textrm{Recall}_c}
\end{equation}
Where c represents each class, while $N$ is the total number of data
and $N_c$ is the number of data with label c. Moreover, we report
the precision, recall and accuracy values separately and compare our
results against the state of the art methods.

We also provide the confusion matrix as a qualitative measure for
both of the datasets. This matrix allows us to interpret how the
model wrongly classifies each of the categories. In the provided
matrices, the rows correspond to the actual labels while columns
represent the predicted classes. We will also take a look at the
directionality of this matrix and discuss how easy it is for a model
to confuse one class with the other but not the other way around.
This approach will allow us to get a look at the specificity of the
classes.

\section{Results and Discussion}
We used the Optuna~\cite{optuna} library to optimize the
hyperparameters of the network. Due to a limited amount of
computational power, we only used this library to search for
iteration number(r), soft-updating value($\eta$) and the initial
learning rate. Twenty trials were conducted while each trial lasted
200 epochs. The best set of hyperparameters were chosen based on the
average validation loss. Moreover, an exponential learning rate
scheduler with a multiplicative factor equal to 0.98 was used to
train the network on both datasets.

\subsection{Results on PAMAP2}
Loss margins were set to 0.95 and 0.05 for this dataset. Moreover,
the iteration number and the soft updating coefficient were set to 3
and 0.1, respectively, while the training proceeded with a batch
size of 64. The test accuracy of our model ranges from 89.18\% to
90.51\% across multiple training sessions of the same model and the
best single epoch based on the validation loss achieves an accuracy
score of 90.51\% on the test set. Due to the observed variance in
the performance of each model during training, we also formed a
horizontal voting ensemble~\cite{ensemble}, based on epochs of a
single model using the top validation scores but the metrics did not
improve when using this ensemble. Table~\ref{pamap2comp} provides a
comparison between the obtained results of our model and the
reported metrics from the state of the art models.
\begin{table}[ht]
    \centering
    \caption{Results on PAMAP2}
    \label{pamap2comp}
    \begin{tabular}[t]{l>{\raggedright}p
    {0.1\linewidth}>{\raggedright\arraybackslash}p
    {0.1\linewidth}>{\raggedright\arraybackslash}p
    {0.1\linewidth}>{\raggedright\arraybackslash}p{0.1\linewidth}}
        \toprule
        &Precision&Recall&wF1 Score&Accuracy\\
        \midrule
        CNN-EF&85.51\%&84.53\%&84.57\%&84.53\%\\
        DeepConvLSTM&87.75\%&86.78\%&86.83\%&86.78\%\\
        PerceptionNet&89.76\%&88.57\%&88.74\%&88.56\%\\
        ARC-Net&\textbf{91.77\%}&\textbf{90.52\%}&\textbf{90.76\%}
        &\textbf{90.51\%}\\
        \bottomrule
    \end{tabular}
\end{table}%
As it can be seen in Table~\ref{pamap2comp}, the results from our
model surpass the state of the art results on all metrics. The
largest gap between ARC-Net and PerceptionNet can be seen in the
precision achieved on the test results which is about 2.01\%.
Furthermore, A consistent improvement is seen on other metrics.
\begin{figure}[t]
    \includegraphics[scale=0.345]{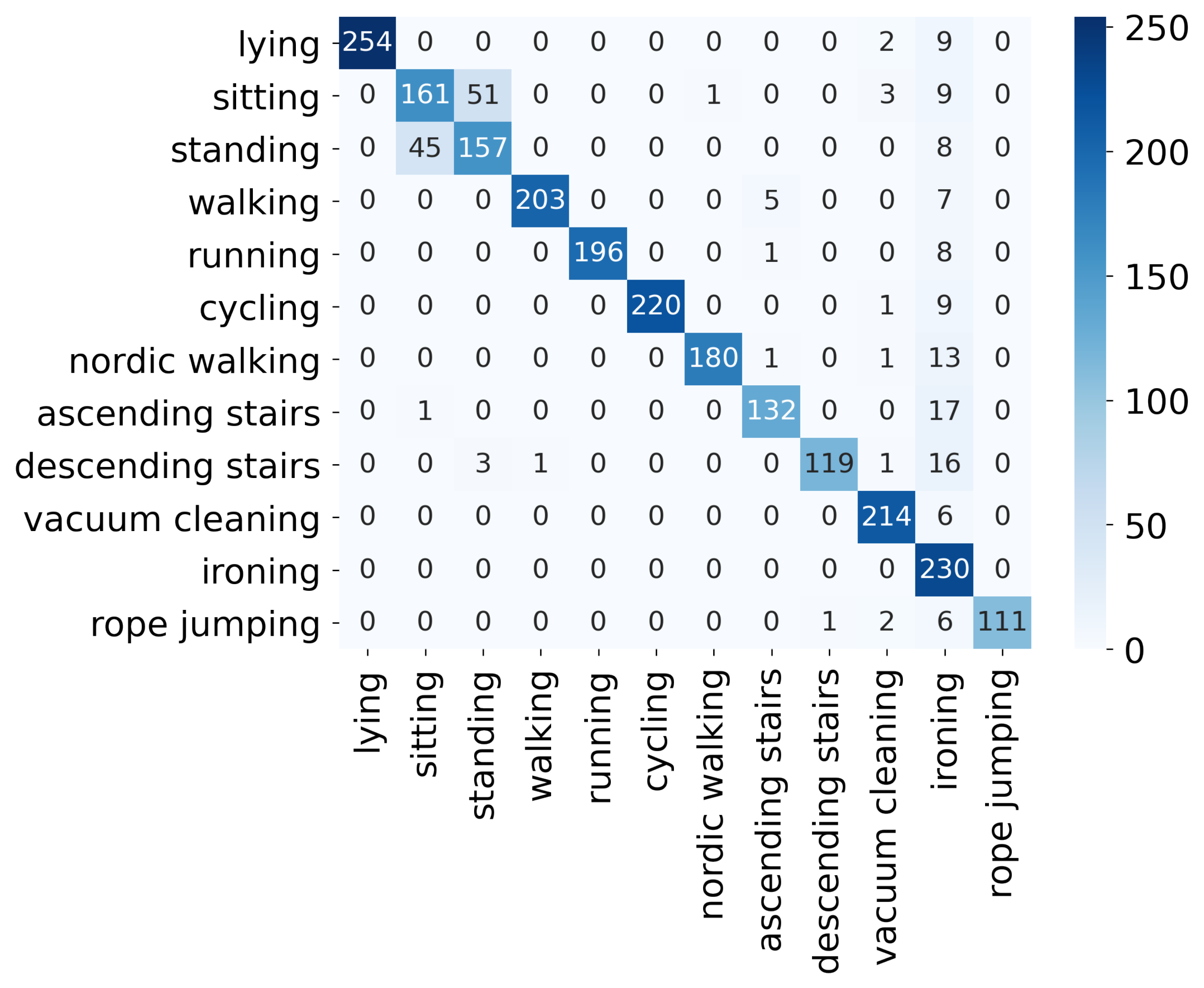}
    \caption{Confusion map of the test set on the PAMAP2 dataset}
    \label{pamap2confmap}
\end{figure}

The  confusion matrix of the test results of the network on the
PAMAP2 dataset is provided in Fig.~\ref{pamap2confmap}. The
directionality of the confusion of the network in classifying
sitting and standing activities suggests that the two activities are
similar to each other based on the provided data. This is expected
since this dataset does not provide any data from an IMU that is
positioned in such a way that would allow an easier differentiation
between the two classes. If an IMU that is placed inside a pocket
has been used, it would allow the network to be able to
differentiate between the two more easily. Furthermore, lying,
cycling, rope jumping and running have a precision of 100\%. The
precision of the ironing activity is low while having a high recall,
specifically ironing has a recall of 100\% while its precision is
only 68.04\%.

\subsection{Results on RealWorld}
The accuracy of our method on the test set ranges from 95.47\% to
95.64\% across multiple training sessions. We report our best epoch
in Table~\ref{realworldcomp} alongside a full comparison of our
network against the state of the art methods. Due to the observed
variance in the performance of the network during training, we also
formed a horizontal voting ensemble based on the validation loss of
each of the epochs of a single model. This way, we were able to
increase the accuracy of our model to 95.92\%. Same as the results
achieved on PAMAP2, a consistent improvement of all the metrics is
seen in Table \ref{realworldcomp}.
\begin{table}[bbb]
    \centering
    \caption{Results on RealWorld}
    \label{realworldcomp}
    \begin{tabular}[t]{l>{\raggedright}p{0.1\linewidth}>{\raggedright\arraybackslash}p{0.1\linewidth}>{\raggedright\arraybackslash}p{0.1\linewidth}>{\raggedright\arraybackslash}p{0.1\linewidth}}
        \toprule
        &Precision&Recall&wF1 Score&Accuracy\\
        \midrule
        DeepConvLSTM&92.83\%&92.65\%&92.63\%&92.65\%\\
        PerceptionNet&94.78\%&94.20\%&94.27\%&94.20\%\\
        ARC-Net&\textbf{96.08\%}&\textbf{95.64\%}&\textbf{95.67\%}&\textbf{95.64\%}\\
        \bottomrule
    \end{tabular}
\end{table}%
Because of the relatively large number of sensors in this dataset,
the number of iterations was increased to 7 while the soft updating
coefficient was dropped to 0.01. By this means, the larger amount of
iterations would allow the network to focus more on the routing of
each primary capsule from each IMU to the high level capsules.
Moreover, the smaller soft updating coefficient would prevent
overrouting of the network.
\begin{figure}[t]
    \hspace{0.15cm}
    \includegraphics[scale=0.30]{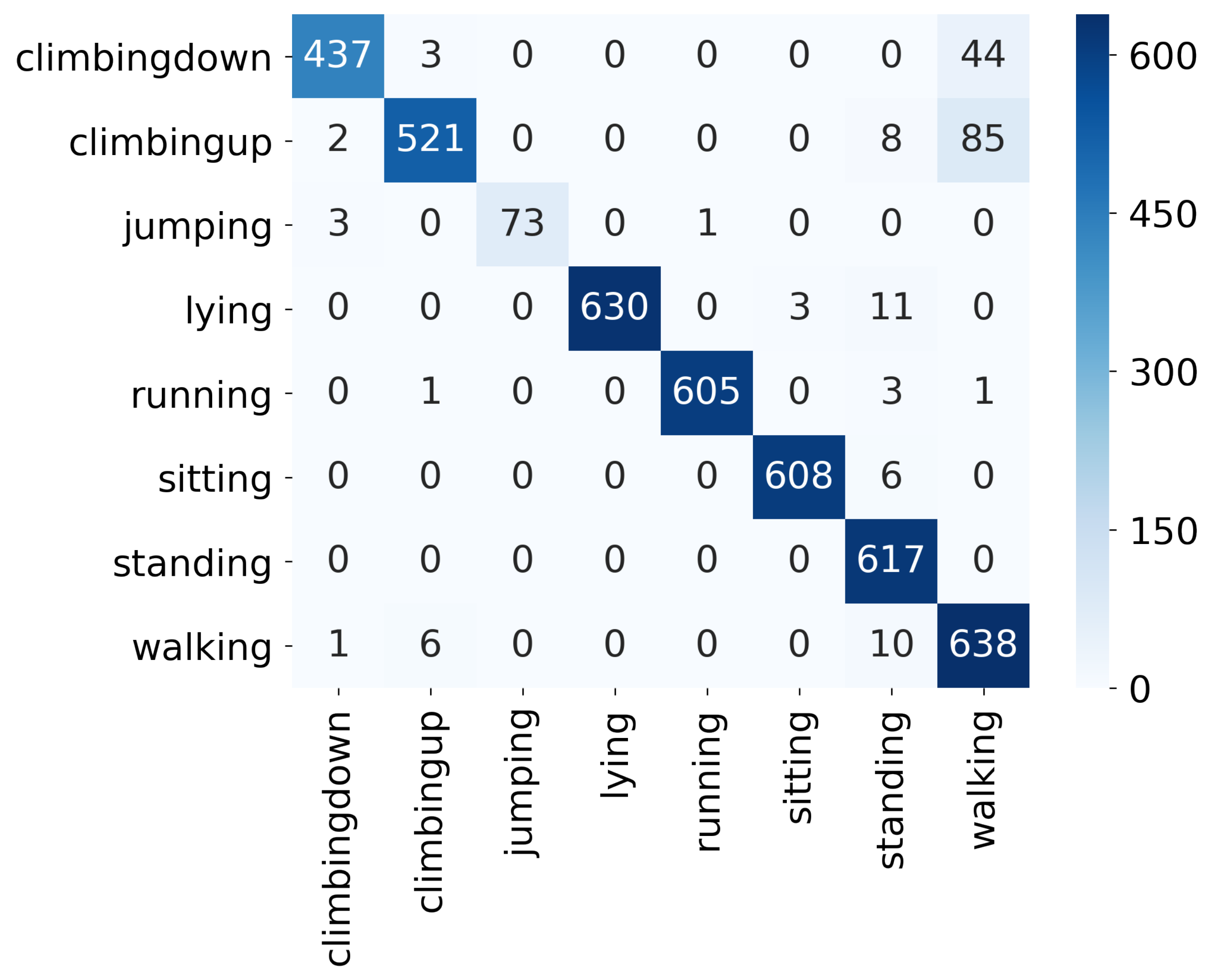}
    \caption{Confusion map of the test set on the RealWorld dataset}
    \label{realworldconfmap}
\end{figure}
The confusion matrix corresponding to the test results of this
dataset is provided in Fig.~\ref{realworldconfmap}.  This
visualization shows that some samples from each activity are
consistently being misclassified as the standing activity. This is
due to the data collection protocol where the subjects are asked to
stand up before performing any of the activities available in the
dataset. Moreover, climbing up and down is also being largely
misclassified as walking. Based on the footage of the activities
provided by~\cite{realworld}, the misclassifications have to do with
the sections of the path where no climbing up or down is performed
and the subject is in transition between two staircases. Therefore,
a large number of these samples are because of the wrong annotations
in the dataset rather than being wrong predictions of the network.

\begin{table}[ttt]
    \centering
    \caption{Changes in Accuracy and F1-score with Modality Corruption}
    \label{noisecomp}
    \begin{tabular}[t]{l>{\raggedright}p{0.1\linewidth}>{\raggedright\arraybackslash}p{0.1\linewidth}>{\raggedright\arraybackslash}p{0.1\linewidth}>{\raggedright\arraybackslash}p{0.1\linewidth}}
        \toprule
        &\multicolumn{2}{c}{PAMAP2}&\multicolumn{2}{c}{RealWorld}\\
        &$\Delta$wF1&$\Delta$Acc&$\Delta$wF1&$\Delta$Acc\\
        \midrule
        PerceptionNet&20.47\%&18.63\%&8.21\%&8.1\%\\
        ARC-Net&\textbf{10.60\%}&\textbf{10.93\%}&\textbf{3.67}\%&\textbf{3.66}\%\\
        \bottomrule
    \end{tabular}
\end{table}%
\begin{figure}[b]
    \hspace{0.4cm}
    \includegraphics[scale=0.346]{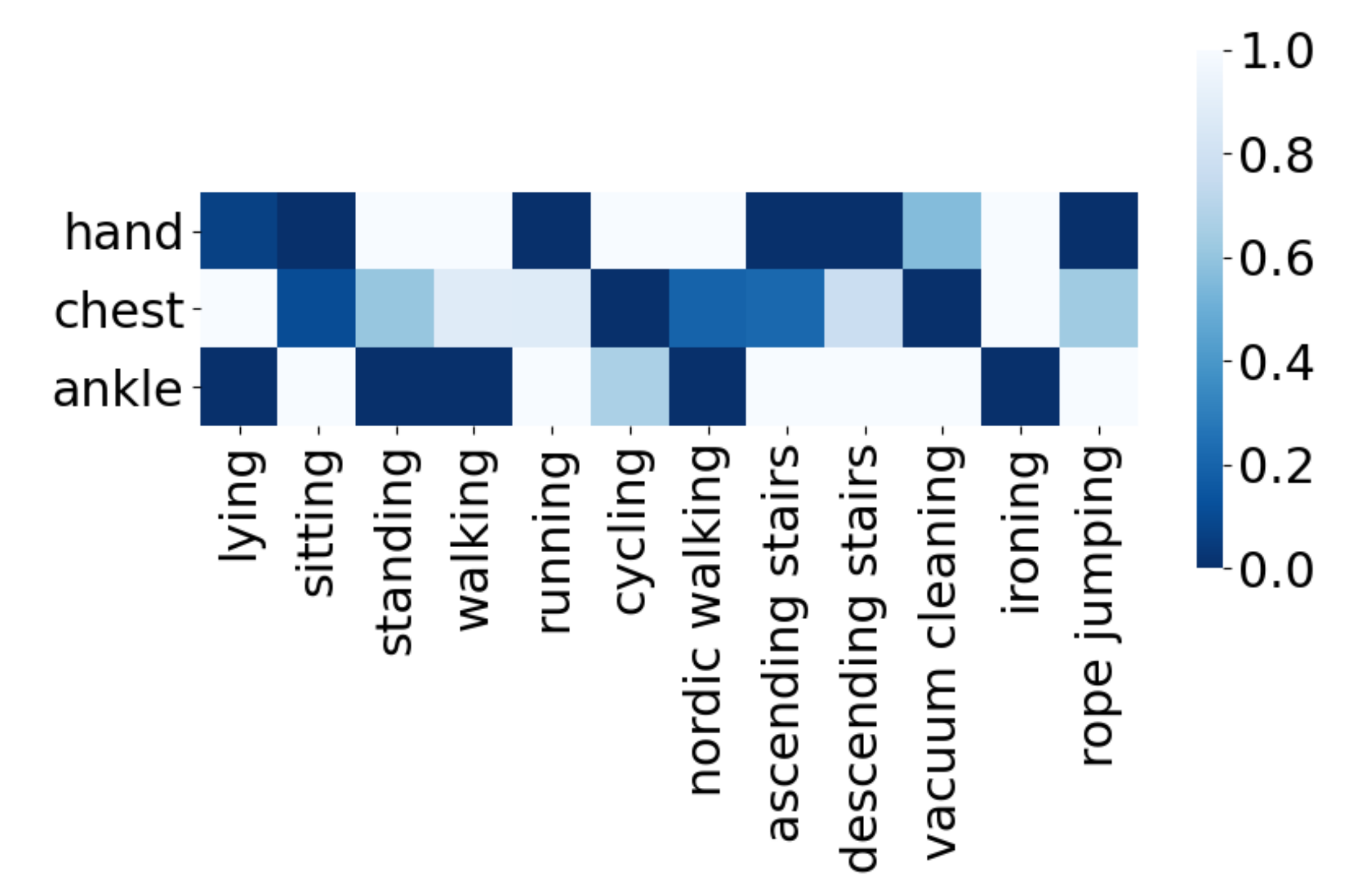}
    \caption{Normalized prior matrix heatmap of the network trained on PAMAP2}
    \label{priormappamap2}
\end{figure}

\subsection{Prior Matrix Visualization}
\begin{figure}[t]
    \hspace{0.15cm}
    \includegraphics[scale=0.37]{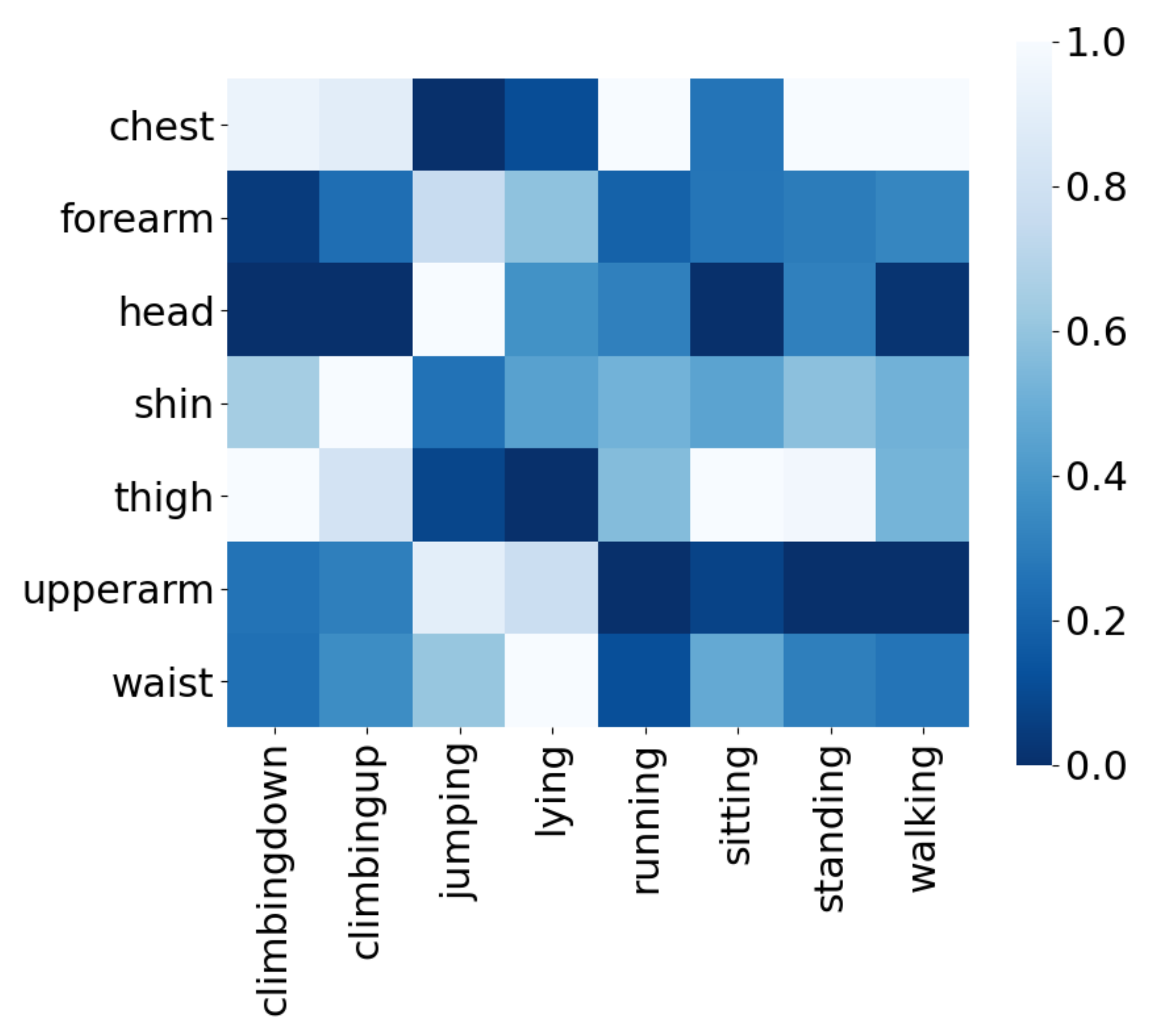}
    \caption{Normalized prior matrix heatmap of the network trained on RealWorld}
    \label{priormaprealworld}
\end{figure}

The prior matrix denoted by $b$ in the routing mechanism of CapsNet
is a learnable parameter and is also responsible for setting the
values that describe the model's prior belief regarding the routing
between two layers of capsules. Therefore, it is possible to extract
the values of this matrix after training and visualize the routing
between each IMU and the performed activities.
Fig.~\ref{priormappamap2} and Fig.~\ref{priormaprealworld} visualize
the prior matrix of our trained network on PAMAP2 and RealWorld,
respectively.

Based on Fig.~\ref{priormappamap2}, the routing that the network has
come up with seems intuitive in activities such as the ones related
to walking or climbing up and down the stairs where the network
relies on the movement of the ankle or hand more than other
modalities to make predictions. Moreover, correlations can be seen
between the heatmaps of PAMAP2 and RealWorld for the same
activities, e.g. the activity of lying relies on the measurements
from the waist/chest when trained on either datasets. Based on
Fig.~\ref{priormaprealworld}, the addition of a thigh modality has
allowed the standing and sitting activities to be more
distinguishable with respect to PAMAP2. This has resulted in a
substantial improvement in the directionality between the two
classes in Fig.~\ref{realworldconfmap} with respect to
Fig.~\ref{pamap2confmap}.

\subsection{Modality Corruption Test}
One of the modalities was randomly corrupted by replacing its
measurements with a zero matrix to simulate the potential case of
modality failure during inference. The drops in the accuracy and
weighted F1 scores of our method and PerceptionNet on the test sets
of each dataset are reported in Table~\ref{noisecomp}. It can be
seen that on both datasets, our approach is significantly more
robust against modality corruptions compared to a CNN only approach.

\section{Conclusions}
In  this paper, we developed a method for human activity recognition
that relies on CapsNet to fuse the information from multiple IMUs.
We tested our approach on two datasets with varying sensor positions
and compared our results against the SOTA. Our results surpassed
that of SOTA by about 2\% in accuracy and weighted F1 score.
Moreover, the confusion matrices of the test set of each dataset
were presented and specificity of the classes was investigated. Our
approach allowed for the visualization of routing between modalities
and activities. Through these visualizations, we were able to
interpret the importance of each modality for correct classification
of each activity. Finally, modality corruption was simulated by
passing an array of zeros instead of one random modality during
inference. Through this test, the capability of our method in noise
rejection was shown. In our approach, it is possible to add extra
modalities without a need for significant changes in the structure
of the network which allows for easier transfer learning. Moreover,
our encoder can be trained on data from various positions of the
body to achieve a general feature extractor.

\end{document}